\RequirePackage{fix-cm}
\let\origvec\vec
\documentclass{svjour3}                     
\smartqed  
\usepackage{graphicx}
\let\vec\origvec
\let\vec\origvec
\usepackage[misc]{ifsym} 
\usepackage{amsmath}
\usepackage[linesnumbered,ruled,vlined]{algorithm2e}
\usepackage{amssymb}
\usepackage{cite}

\begin{document}

\title{Solving Large-Scale Multi-Objective Optimization via Probabilistic Prediction Model
\thanks{Min Jiang is the corresponding author.\\This work was supported by the National Natural Science Foundation of China (No.61673328).}
}

\titlerunning{LT-PPM}        

\author{Haokai Hong  \textsuperscript{1} \and
        Kai Ye       \textsuperscript{1}  \and \\
        Min Jiang    \textsuperscript{1,2}$^{(\textrm{\Letter})}$ \and
        Donglin Cao    \textsuperscript{1,3}$^{(\textrm{\Letter})}$ \and
        Kay Chen Tan \textsuperscript{4}
}

\authorrunning{H. Hong et al.} 

\institute{
    {1} Department of Artificial Intelligence, Key Laboratory of Digital Protection and Intelligent Processing of Intangible Cultural Heritage of Fujian and Taiwan, Ministry of Culture and Tourism, School of Informatics, Xiamen University, Fujian, China, 361005.\\
    {2}  \email{minjiang@xmu.edu.cn}\\ 
    {3}  \email{another@xmu.edu.cn}\\ 
    {4} Department of Computing, Hong Kong Polytechnic University, Hong Kong. \email{kaychen.tan@polyu.edu.hk} \\ 
}

\date{Received: date / Accepted: date}

\maketitle

\begin{abstract}

The main feature of large-scale multi-objective optimization problems (LSMOP) is to optimize multiple conflicting objectives while considering thousands of decision variables at the same time. An efficient LSMOP algorithm should have the ability to escape the local optimal solution from the huge search space and find the global optimal. Most of the current researches focus on how to deal with decision variables. However, due to the large number of decision variables, it is easy to lead to high computational cost. Maintaining the diversity of the population is one of the effective ways to improve search efficiency. In this paper, we propose a probabilistic prediction model based on trend prediction model and generating-filtering strategy, called LT-PPM, to tackle the LSMOP. The proposed method enhances the diversity of the population through importance sampling. At the same time, due to the adoption of an individual-based evolution mechanism, the computational cost of the proposed method is independent of the number of decision variables, thus avoiding the problem of exponential growth of the search space.  We compared the proposed algorithm with several state-of-the-art algorithms for different benchmark functions. The experimental results and complexity analysis have demonstrated that the proposed algorithm has significant improvement in terms of its performance and computational efficiency in large-scale multi-objective optimization.

\end{abstract}

\section{Introduction}
\label{sec:intro}
Many optimization problems in the real world involve hundreds or even thousands of decision variables and multiple conflicting goals. These problems are defined as large-scale multi-objective optimization problems (LSMOP) \cite{RN104, RN103}. Large-scale multi-objective optimization problems have emerged from many real-world applications \cite{RN231}. For example, in the problem of designing protein molecular structure, thousands of decision variables are considered. Besides, in a routing system with large number of network nodes, the distribution of transmission capacity of each node is regarded as a decision variable, whereas the energy consumption, robustness and stability of the network system constitute a large-scale multi-objectives optimization problem \cite{7533424}. Since the size of the search space has an exponential relationship with the number of decision variables, which leads to the problem of dimensionality explosion in LSMOP \cite{RN105}, finding the optimal solution in LSMOP is more challenging than ordinary multi-objective optimization problems(MOP) \cite{5415586}. Therefore, an excellent large-scale multi-objective optimization algorithm (LSMOA) should overcome this problem through searching in the decision space effectively, which is essential for solving LSMOP.

The fact that LSMOP is more difficult to solve than MOP with a small number of decision variables makes it challenging for the multi-objective evolutionary algorithm (MOEA) to effectively explore the search space. Generally speaking, MOEA is tend to converge to a local optimum prematurely, or it may converge to a relatively large area\cite{RN121}. Therefore, it has been proved through experience that most existing MOEAs cannot solve LSMOPs \cite{RN122, RN82}. In fact, most existing MOEAs improve the effectiveness of solving MOP by enhancing environmental selection (that is, the strategy of selecting solutions for the next generation from the current population) \cite{RN122, RN124, RN125}. However, the search efficiency is not sufficient for finding the optimal solution in a large search space. Thus, it is important to find a LSMOA that can search for the optimal solution in the space of LSMOPs.

In order to solve LSMOP, various LSMOAs have been presented in recent years \cite{2008Large, 8720021, 9047876, 8765790}. These proposed LSMOAs can be roughly divided into the following three categories. The first category of LSMOAs is benefits from the decision variable analysis. A large-scale evolutionary algorithm based on the decision variable clustering (LMEA) \cite{RN82} is representative in this category, which divides the decision variables into two categories via clustering method. Then, by focusing on convergence and diversity respectively, two different strategies are used to optimize the variables related to convergence and diversity. The second category of LSMOAs is based on the decision variable grouping. This type of LSMOA divides the decision variables into several groups, and then optimizes each group of decision variables. For example, C-CGDE3 \cite{RN84} maintains several subsets of equal-length decision variables obtained by variable grouping as subpopulations. The third category is based on problem conversion. A representative framework of this category is called Large-scale Multi-Objective Optimization Framework (LSMOF) \cite{RN88}. In LSMOF, the original LSMOP is converted into a low-dimensional single-objective optimization problem with some direction vectors and weight variables, the method aimed to guide the population towards the optimal solution.

In the methods of decision variable analysis and decision variable grouping, due to that fact that the algorithm relies on the research on decision variables, the algorithm complexity is relatively high. However solving LSMOP should not only consider how to escape from local optimum, but also how to reduce the impact of the number of decision variables on the complexity of the algorithm \cite{10.5555/3298239.3298367}. Inspired by the advantages of Trend Prediction Model (TPM), we found that the Trend Prediction Model can be used to predict the optimal solution. The TPM is used to estimate and predict the probability of potential particles at each position in the next generation, thereby increasing the uncertainty of particle movement and achieving better prediction results. With the advantages of trend prediction model (TPM), the method not only maintains the diversity of the population \cite{9122031}, but also keeps the computational complexity independent of decision variables. In the specific implementation, TPM was applied to sample individuals from the parent and expand into the offspring population.

In this paper, we integrate the Trend Prediction Model into a probabilistic prediction model to obtain a novel LSMOA algorithm. This method is a probabilistic prediction model for solving large-scale multi-objective optimization based on trend prediction model (LT-PPM). Our proposed algorithm selects individuals and expands them from the previous generation to achieve the purpose of expanding and generating a new offspring population via TPM, which ensures the diversity of the offspring population. The contributions of this work are summarized as follows.

 \begin{enumerate}
  \item The proposed algorithm framework is an effective tool to solve the problem of large decision variable space and overcome the tendency to fall into a local optimum in LSMOP. Our method uses the trend prediction model to maintain the quality and diversity of the population simultaneously to escape from the local optimum, which provides a novel idea to solve LSMOP.
  \item Taking into account the high dimensionality of the decision variables of the LSMOP problem, the complexity of our proposed method is not directly related to the scale of the decision variables, but only related to the number of objectives and populations. While escaping from the local optimum, the influence of large-scale decision variables on the computational complexity is also eliminated in our method.
\end{enumerate}


\section{Preliminary Studies and Related Works}
\label{sec:prestu}

In this section, the definition of large-scale multi-objective optimization problems is presented. And then existing methods to solve LSMOPs are introduced.

\subsection{Large-scale Multi-objective Optimization}
\label{sec:LSMOP}

The mathematical form of LSMOPs is as follows:
\begin{equation}
\begin{aligned}
\label{equ:LSMOP}
\mathbf{Minimize}\ F(x) = & \left \langle f_1(x),f_2(x),...,f_m(x) \right \rangle \\
& s.t. \ x\in \Omega
\end{aligned}
\end{equation}
where $x = (x_1, x_2, \dots, x_n)$ is the $n$-dimensional decision vector,\\
$F = (f_1(x),f_2(x),...,f_m(x))$ is the $m$-dimensional objective vector.

It should be noticed that the dimension of decision vector $n$ is more than 100 in LSMOP \cite{RN85}. When dealing with multi-objective optimization problems, evolutionary algorithms (EAs) are usually based on individual evolution strategies, genetic operators are used to evolve individuals, so that EAs have good global convergence and versatility. The evolutionary framework we use in this paper is an emerging branch in this field called the Probabilistic Prediction Model (PPM) \cite{JIANG2018203, 8285444}, PPM promots the evolution of the entire population by establishing a probability model on the Pareto-optimal Front.

\subsection{Related Works}
\label{sec:RW}
In this part, we will introduce some state-of-art large-scale multi-objective optimization algorithms. Much progress has been made in solving LSMOPs. Most of existing methods can be divided into three categories: based on decision variable analysis, based on decision variable grouping and based on problem transformation.

The first category is based on the analysis of decision variables. Zhang et al. \cite{RN82} proposed a large-scale evolutionary algorithm (LMEA) based on clustering of decision variables. Another representative of this type of algorithm is MOEA/DVA \cite{RN106}, in which the original LSMOP is decomposed into many simpler sub-MOPs. Then, the decision variables in each sub-problem are optimized as an independent sub-component. The disadvantage of decision variable analysis-based LSMOAs is that the connection between decision variables could be incorrectly determined to update the solution, which may lead to local optimization \cite{RN82}. To solve large-scale MOPs that pose a great challenge to the balance between diversification and convergence by considering a correlation between position and distance functions, a novel indicator-based algorithm with an enhanced diversification mechanism was developed by Hong et al. \cite{8533425}.

The second type of LSMOAs is decision variable grouping-based framework \cite{RN83}. The method proposed by Antonio et al. \cite{RN84} maintained several independent subgroups. Each subgroup is a subset of the same-length decision variables obtained by variable grouping (for example, random grouping, linear grouping, ordered grouping or differential grouping). All sub-populations work together to optimize LSMOP in a divide-and-conquer manner. Zille et al. \cite{RN89} proposed a weighted optimization framework (WOF) for solving LSMOP. Decision variables are divided into many groups, and each group is assigned a weight variable. Then, in the same group, the weight variable is regarded as a sub-problem of a sub-space from the original decision space. Liu et al. \cite{LIU2020100684} proposed an algorithm in which the random dynamic grouping was used to substitute ordered grouping to improve the performance of the WOF framework.

In this category, there is a novel idea that based on competitive optimizer. Tian et al. \cite{RN85} proposed LMOCSO based on a competitive population optimizer to solve LSMOP. The method used a two-stage strategy to update the position of the particles. Another idea of this category is based on collaboration. Bergh et al. \cite{RN87} proposed a method to transform particle swarm optimization algorithm into a collaborative framework. It divides the search space into low-dimensional subspaces to overcome the exponential increase in difficulty and ensures that it can search every possible area of the search space. And Shang et al. \cite{RN86} proposed the idea of cooperative co-evolution, which was used to deal with large-scale multi-target capability routing problems. Another cooperative coevolutionary algorithm was proposed by Li et al. \cite{10.1145/3205651.3208250}, a fast interdependency identification grouping method was utilized for large-scale multi-objective problems.

However, for LSMOA based on the grouping of decision variables, two related variables can be divided into different groups, which may lead to local optimization \cite{RN83}.

The third category is based on problem transformation. In order to enhance the pressure of convergence, one intuitive solution is to directly convert the definition of traditional Pareto dominance, such as $\theta$-dominance \cite{RN116}, $L$-optimality \cite{RN113}, $\epsilon$-dominance \cite{RN112}, preference order ranking \cite{RN115} and fuzzy dominance \cite{RN114}. To enhance the convergence performance of evolutionary algorithm in solving large-scale multi-objective problems, Qin et al. \cite{9367299} proposed a method for generating guiding solutions along search directions defined by a few chosen individuals in the current parent population. Another kind of idea that belongs to this category is to combine traditional dominance with other convergence-related metrics. MOEAs that fall into this category include Grid-based Evolutionary Algorithm (GrEA) \cite{RN118}, NSGA-II based on substitute distance assignment \cite{RN117}, many-Objective Evolutionary Algorithm based on Directional Diversity and Favorable Convergence (MaOEA-DDFC) \cite{RN120}, Preference-Inspired Co-Evolutionary Algorithm (PICEA-g) \cite{RN119}, and Knee point driven Evolutionary Algorithm (KnEA) \cite{RN109}.

In addition to the above-mentioned convergence-related problem transformation, for LSMOP, He et al. \cite{RN88} introduced a general framework called the large-scale multi-objective optimization framework (LSMOF). LSMOF reconstructs the original LSMOP into a low-dimensional single-objective optimization problem with some direction vectors and weight variables, aiming to lead the population to POS. Recently, He et al. \cite{RN90} designed an evolutionary multi-objective optimization (GMOEA) driven by GANs, which produces offspring from GANs.

As the number of decision variables increases linearly, the complexity (as well as volume) of the search space will increase exponentially, and the methods of analysis or grouping of decision variables presented above cannot tackle the curse of dimensionality. Therefore, in this work, we focus on how to use TPM to maintain population diversity to avoid falling into local optimum while avoiding the curse of dimensionality caused by large-scale decision variables to better solve LSMOP.

\section{A Probabilistic Prediction Model Based on Trend Prediction Model}
\label{sec:LTPPM}

\subsection{Overview}
\label{sec:Ov}

Based on the Generating-Filtering strategy, LT-PPM combines the importance sampling method with the non-parametric density estimation function to propose a different selection method for candidate solution. Specifically, LT-PPM uses a kernel density estimation method, a non-parametric probability density estimation method to assign a sparseness to each individual, and then uses importance sampling to select subsequent solutions. This method of selecting candidate solutions aims to ensure the balance between exploitation and exploration in terms of time and space: on the one hand, the sparseness will change with generation, on the other hand, sparseness itself will also guide the method to achieve the balance in the large-scale search space.

We first give the definition of particles, and then in the remainder of this chapter we will introduce importance sampling based on kernel density estimation, trend prediction model, LT-PPM and the complexity analysis of the algorithm.

\begin{definition}[Particle]
\label{def:particle}
    \\ For an optimization problem with $n$ optimization goals, the particle is a two-tuple $p =(x, \boldsymbol{v})$, where $x$ represents a solution of the multi-objective optimization problem, and $\boldsymbol{v}$ is a unit vector in $n$-dimensional space representing the direction of the particle. 
\end{definition}

Note that the direction vector of the first generation population is randomly initialized. In the subsequent offspring population, the direction vector of each individual is the direction of movement when their parent particle produce them.

\subsection{Importance Sampling Based on Kernel Density Estimation}
\label{sec:IS}
Importance Sampling is defined as follows,
\begin{definition}[Importance Sampling]
\label{def:IS}
    \\ When calculating the expectation of the statistic $f(x)$ on the original probability distribution $p(x)$, one can calculate the expectation of $f(x)\cdot p(x)/q(s)$ on a certain importance probability distribution $q(x)$,
    \begin{equation}
    \begin{aligned}
      & \mathbf E _p[f(x)] \\
    = & \int _x p(x) \cdot f(x) dx \\
    = & \int _x q(x) \cdot f(x) \frac {p(x)} {q(x)} dx \\
    = & \mathbf E _q [f(x)\frac {p(x)} {q(x)}]
    \end{aligned}
    \end{equation}
\end{definition}

Assuming that the current population is \\ $P = \{x_1, \dots, x_n\}$, and corresponding particles are $P' = \{p_1, \dots, p_n\}$. The probability density of the population in different regions reflects the degree of density in the search space: where the probability density is high, the population is denser. Therefore, the density of particle $p_k$ can be defined as its kernel density estimate,
\begin{equation}
\begin{aligned}
\label{equ:Density}
\mathbf{Density}(p_k) = \frac 1 {nh} \sum_{i=1}^n \kappa(\frac {p_k-p_i} h)
\end{aligned}
\end{equation}
where $\kappa(\cdot)$ is the kernel function, and the parameter $h$ is used to adjust the width of the kernel density curve: when $h$ is larger, the distribution curve is smoother, when $h$ is smaller, the distribution curve is steeper.

The \textbf{Sparseness} of particle $p_k$ is defined as the reciprocal of \textbf{Density},
\begin{equation}
\begin{aligned}
\label{equ:Sparseness}
\mathbf{Sparseness}(p_k) = \frac 1 {\mathbf{Density}(p_k)}
\end{aligned}
\end{equation}

The probability density of the population in different regions reflects the degree of density in the search space: where the probability density is high, the population is denser. Therefore, the density of particle $p_k$ can be defined as its kernel density estimate, The \textbf{Sparseness} of particle $p_k$ is defined as the reciprocal of \textbf{Density}. Based on the \textbf{Sparseness} of each particle, Importance Sampling for our method can be defined. Since each particle is randomly selected, particle $p_k$ obeys a uniform distribution $U(p_k) = 1/n$. Then we have importance distribution,
\begin{equation}
\begin{aligned}
\label{equ:ISP}
IS(p_k) = \frac {\mathbf{sparseness}(p_k)} {\sum_i\mathbf{sparseness}(p_i)}
\end{aligned}
\end{equation}

The number of offspring that the particle $p_k$ will produce is proportional to $S(p_k)$. The expectation value of sparseness describes the sampling process, and its calculation formula is: $E_U[S(p_k)]$.
However, in MOPs, it is not easy to determine how many offspring the selected particle should produce, so we do not want to directly use this method for sampling. The original sampling process can be rewritten as,
\begin{equation}
\begin{aligned}
  & \mathbf E_U[S(p_k)] =  \mathbf E_S[S(p_k) \cdot \omega(p_k)] = \mathbf E_S[U(p_k)] \\
\end{aligned}
\end{equation}
where $\omega(p_k) = \frac {U(p_k)} {S(p_k)}$.

This means that we can select particles by the importance distribution $S$, and then produce the same number of offspring. Algorithm \ref{alg:IS} summarizes the above sampling process. In each iteration, first, the sparseness of each particle is calculated, and then the importance distribution $S$ is constructed, then an individual is sampled on the distribution $S$, and the individual and its velocity are returned. This sampling method does not rely on the analysis of decision variables, avoiding the influence of large-scale decision variables on computational complexity.

\begin{algorithm}
    \caption{ISample($POS$)}
    \label{alg:IS}
    \KwIn{
        $POS$ (Pareto-optimal Set), $h$ (bandwidth)
    }
    \KwOut{$p_i$ (sampled particle), $x_k$(solution of $p_i$), $\boldsymbol{v_k}$(velocity of $p_i$)}
    $n =$ size of $POS$ \;
    \For{$p_i$ in $POS$}{
        $\mathbf{Density}(p_k) = \frac 1 {nh} \sum_{i=1}^n \kappa(\frac {p_k-p_i} h)$ \;
        $\mathbf{Sparseness}(p_k) = \frac 1 {\mathbf{Density}(p_k)}$ \;
    }
    \For{$x_i$ in $POS$}{
        Constructe $IS(x_i)$ according to Equation \ref{equ:ISP} \;
    }
    Sampled $x_k$ based on $IS(x_i)$ \;
    \Return $x_k$, $\boldsymbol{v_k}$
\end{algorithm}

\subsection{Trend Prediction Model}
\label{sec:TPM}
After selecting the particles to be expanded via above selection process, it is necessary to sample on a specific distribution represented by selected particles to generate new individuals. Physically, if an object is not affected by external forces, it will maintain its original state of motion. Therefore, if a particle moves in the direction $\boldsymbol{v}$, the displacement direction $\boldsymbol{s}$ of the particle at the next moment is equal to $\boldsymbol{v}$.

In order to effectively move to the POF, the randomness is added to the movement of the particle, so that the direction of movement only provides information on the trend of the movement instead of directly determining the place at the next moment. Such trend information can well guide the particles to update, thereby approaching the POF faster and more accurately.

\begin{definition}[Trend Prediction Model, TPM]
\label{def:TPM}
    \\ For any given direction vector $\boldsymbol{v} \in \mathbb{R}^n$ and distance measurement function $d(\cdot, \cdot)$. A probability distribution $P$ on $D\subset \mathbb{R}^n$ is called as a trend prediction model (TPM).
\end{definition}

For the two vectors $\boldsymbol{t_1}$ and $\boldsymbol{t_2}$ in the model, if it satisfies:
\begin{equation}
\begin{aligned}
\forall \boldsymbol{t_1}, \boldsymbol{t_2} \in D, d(\boldsymbol{v}, \boldsymbol{t_1}) \leq d(\boldsymbol{v}, \boldsymbol{t_2}) \to P(\boldsymbol{t_1}) \geq P(\boldsymbol{t_2})
\end{aligned}
\end{equation}
then it means that the vector $t_1$ is closer to the direction vector $v$ than $t_2$, and $t_1$ should have a higher probability of occurrence.

\begin{algorithm}
    \caption{randpick($\boldsymbol{v}, \theta$)}
    \label{alg:rp}
    \KwIn{ $\boldsymbol{v} \in \mathbb{R}^n$ (Unit direction vector) }
    \KwOut{$\boldsymbol{u}$ (Random direction vector)}
    $n = $ Dimension of $\boldsymbol{v}$\;
    $t = $ The subscript of the first non-zero vector in $\boldsymbol{v}$\;
    \For{$i=1,2,\dots,n-1$}{
        \If {$i=t$} {
            $\boldsymbol{\beta_i}$ = The vector that only the $n$th element is $1$ and other elements are $0$;
        }\Else {
            $\boldsymbol{\beta_i}$ = The vector that only the $i$th element is $1$ and other elements are $0$;
        }
    }
    $R$ = Apply $\mathbf{Schmidt\ orthogonalization}$ to the matrix $[\boldsymbol{v}, \boldsymbol{\beta_1},\dots, \boldsymbol{\beta_{n-1}}]$\;
    $\boldsymbol{u}$ = Sampled from the $n-1$ dimensional unit hyperspherical coordinate system \;
    $\boldsymbol{u} = R^{-1} \cdot [\cos(\theta)\boldsymbol{u}]^T$ \;
    \Return $\boldsymbol{u}$
\end{algorithm}

In the trend prediction model, particle movement has a moving trend, which is indicated by the unit vector $\boldsymbol{v}$. The method aims to design the \textbf{TPM} distribution so that the closer the sampling vector to $\boldsymbol{v}$, the higher the sampling probability. For this, $\mathbf{TPM}_h$ is introduced. Given the distance measurement method $d$ is the $\cos$ distance, the domain $D$ is the unit hypersphere, and sampled from the Gaussian distribution with the expectation of 0 and the variance of $1/h$, which is the bandwidth. Since the domain is a unit hypersphere, $\mathbf{TPM}_h$ is actually a probability distribution on a direction vector. In this distribution, directions close to the vector $\boldsymbol{v}$ have a higher probability of being selected (that is, the angle with $\boldsymbol{v}$ is small), otherwise the probability is low.

After sampling an angle $\theta$ on the $\mathbf{TPM}_h$ system, it is necessary to randomly select one among all the vectors whose angle is $\theta$ with $\boldsymbol{v}$. In order to sample all vectors whose angle is $\theta$ with $\boldsymbol{v}$, we can construct a rotation matrix $R$ via Schmidt Orthogonalization to rotate the vector $\boldsymbol{v}$ to the $X$ axis. Then, randomly select a vector on the $n-1$ dimensional hypersphere. Finally, the sampled vector is rotated back to the original coordinate space through the inverse transformation $R^{(-1)}$.

Algorithm \ref{alg:rp} is the process of generating offspring direction vectors according to the parent individual by importance sampling. The complete trend prediction model is given in Algorithm \ref{alg:TPM}, where ISample (POS) is Algorithm \ref{alg:IS}. In the specific strategy of generating offspring, after sampling the direction vector by Algorithm \ref{alg:rp}, it is necessary to continue to determine the particle's movement step length. The actual step length is sampled from a normal distribution with a expectation of 0 and a variance of $h$.

\begin{algorithm}
    \caption{TPM($POS, h, N$) (Trend Prediction Model)}
    \label{alg:TPM}
    \KwIn{
        $POS$ (Pareto-Optimal Set), $h$ (bandwidth), $N$ (number of samples)
    }
    \KwOut{$S$(Sampled set)}
    $S = \emptyset $ \;
    \For{$i=1,2,\dots,N$} {
        $<p, v> = \mathbf{ISample}(POS)$ \;
        $\theta$ = Sampled from $\mathbf{Norm}(0, 1/h)$ \;
        $u = \mathbf{randpick}(v, \theta)$ \;
        $\lambda = $ Sampled from $\mathbf{Norm}(0, h)$ \;
        $\lambda = \lceil \lambda \rceil$ \;
        $S = S \cup \{ p + \lambda u\}$ \;
    }
    \Return $S$ \;
\end{algorithm}

\subsection{Large scale TPM based Probabilistic Prediction Model\ (LT-PPM)}
\label{sec:LT-PPM}
Combining the importance sampling and trend prediction model, the complete algorithm implementation of LT-PPM is given in Algorithm 4.
The key steps of the algorithm are:

\textbf{Generating}: Select the particles to be expanded through importance sampling based on kernel density estimation, and then predict the next generation population through the trend prediction model;

\textbf{Filtering}: Screen out outstanding individuals from the parent population and offspring population.

\begin{algorithm}
    \caption{Large scale TPM based PPM}
    \label{alg:LTPPM}
    \KwIn{
        $N$(size of restricted population), $P$(size of population), $e$ (the maximum evaluations), $r$(attenuation coefficient)
    }
    \KwOut{$POS(the\ \mathbf{POS})$}
    Random initial $P$ population $POS$ \;
    \While {Used evaluations $\leq e$} {
        $S = sample(POS, 1/h, P)$ \;
        Update $POS$ and based on $S$ \;
        Remove the densest individuals repeatedly until $|POS| \leq N$ \;
        $h = h \cdot r $\;
    }
    \Return $POS$ \;
\end{algorithm}

\begin{figure}
\centering
\includegraphics[width=0.9 \textwidth]{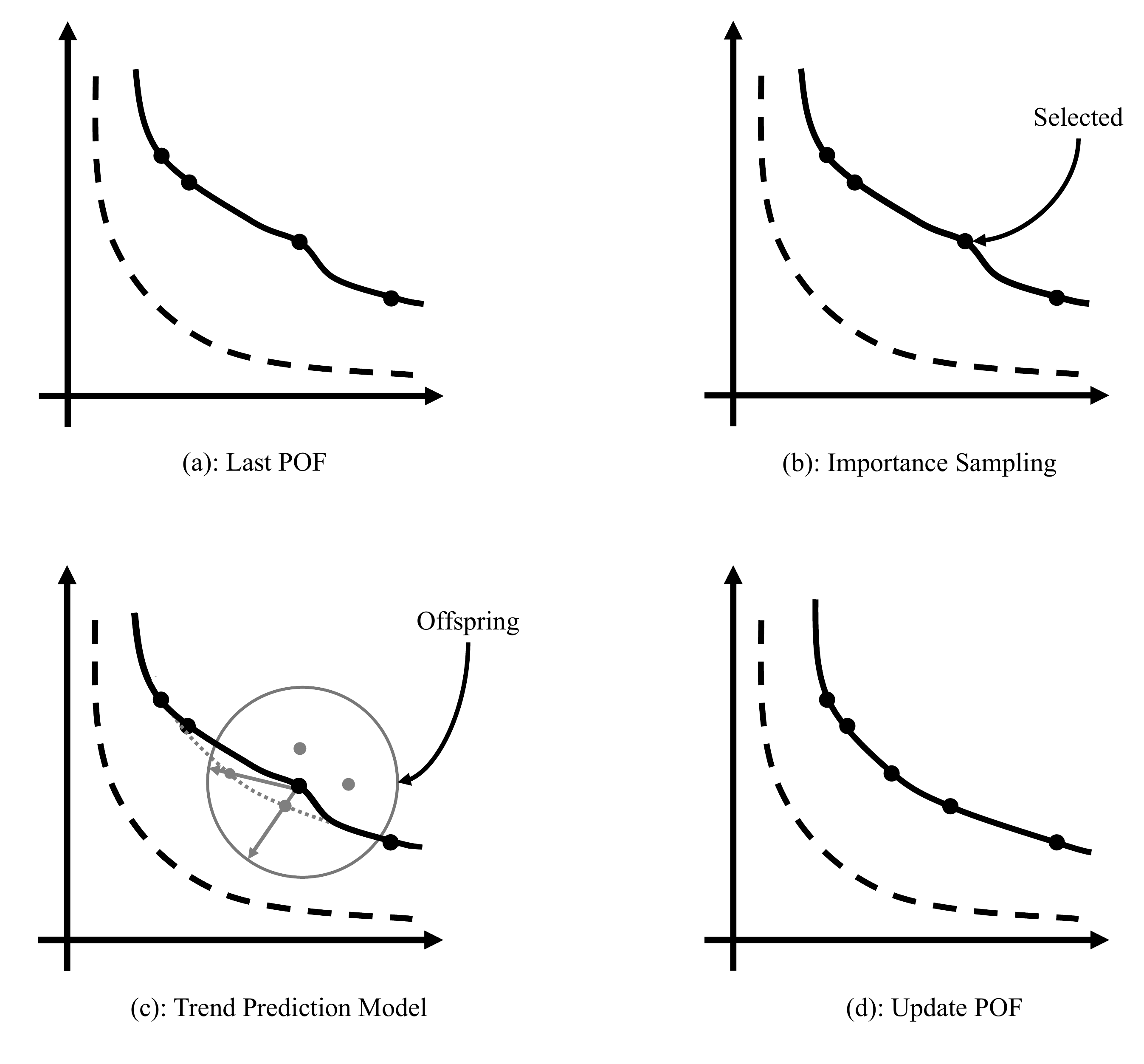}
\caption{Algorithm Framework of LT-PPM}
\label{fig:ltppm}       
\end{figure}

The algorithm framework of LT-PPM is briefly illustrated in conjunction with Figures \ref{fig:ltppm}(a) to \ref{fig:ltppm}(d). In a certain generation, the LT-PPM algorithm maintains a population and calculates the corresponding POS and POF (the solid line in Figure \ref{fig:ltppm}(a) represents the POF of this generation, and the dashed line represents the real POF). Then use the non-parametric kernel density estimation method to estimate the probability density of the distribution corresponding to the POS, and use the importance sampling method to select a particle. The principle is that particles with higher sparseness will have a higher probability of being selected (Figure \ref{fig:ltppm}(b)), while particles in dense particles have a lower probability of being selected. After that, apply the trend prediction model to the selected particles to obtain a series of new solutions (Figure \ref{fig:ltppm}(c), where gray dots represent the newly generated population). Finally, update POS and POF according to the newly generated individuals (Figure \ref{fig:ltppm}(d)).

\subsection{Complexity Analysis}
\label{sec:CA}
Assuming a $n$ objectives LSMOP with $m$ population, then the time to calculate all objectives is $\mathcal{O}(n)$. In our algorithm, $m$ individuals are sampled in each iteration, and each individual requires $\mathcal{O}(m)$ time to sample, therefore the time complexity of the sampling process in each iteration is $\mathcal{O}(m^2)$. Suppose the evaluation is $e$, the actual interation times is $e / m$, so the total complexity of the algorithm is $\mathcal{O}(nme)$.

The complexity of the algorithm is proportional to the square of the number of population and the number of objectives, and has no direct relation with the scale of decision variables of the problem. Therefore, our proposed LT-PPM can avoid the influence of large-scale decision variables on the computational complexity in solving large-scale multi-objective optimization problems.

\section{Experiments and Analysis}
\label{sec:EaA}

\subsection{Algorithms in Comparison and Test Problems}
\label{sec:ACTP}
The proposed LT-PPM was experimented on large-scale multi-objective problems LSMOP1 - LSMOP9 \cite{RN96} and compared with several state-of-the-art algorithms, including LMEA \cite{RN82}, LMOCSO \cite{RN85}, NSGAIII \cite{RN110} and RMMEDA \cite{RN111}. According to \cite{RN92}, the parameters of LMEA, RMMEDA and LMOCSO algorithms are set to default values. For all test problems, the number of optimization objectives is set to 3. The remaining parameters are set as follows,
\begin{enumerate}
  \item Population size: The population size $N$ is set to 300.
  \item Termination condition: The maximum evaluation times $e$ of all compared LSMOAs is set to 100000.
  \item Parameter of LT-PPM: The bandwidth attenuation coefficient $r$ in algorithm \ref{alg:LTPPM} is set to 0.9.
\end{enumerate} 

\subsection{Performance Indicators}
\begin{enumerate}
  \item Schott's Spacing (SP) \cite{RN97}: It measures the uniformity of the solution found by the algorithm. The smaller the SP value, the more uniform the distribution of solutions obtained by the algorithm. It is calculated by the following formula:
  \begin{equation}
  \begin{aligned}
  SP = \sqrt {\frac 1 {n_{\text{POF}^*}-1} \cdot \sum_{i=1}^{n_{\text{POF}^*}} (E_i-\bar E)^2}
  \end{aligned}
  \end{equation}
  where $\text{POF}^*$ is the true POF of the multi-objective optimization problem, $E_i$ represents the Euclidean distance between the $i$-th solution in $\text{POF}^*$ and its closest solution, and $\bar E$ represents the mean value of $E_i$.

  \item Inverted Generational Distance (IGD) \cite{RN98}: IGD is a metric used to quantify the convergence of the solution obtained by the multi-objective optimization algorithm. When the IGD value is small, the convergence of the solution is improved. IGD is defined as
  \begin{equation}
  \begin{aligned}
  IGD(\text{POF}^*, \text{POF}) = \frac 1 n \cdot \sum_{p^*\in \text{POF}^*}\min_{p \in \text{POF}} ||p^*-p||^2
  \end{aligned}
  \end{equation}
  where $\text{POF}^*$ is the true POF of the multi-objective optimization problem, POF is the approximate set of POF obtained by the multi-objective optimization algorithm, and $n$ is the number of solutions in $\text{POF}^*$.
\end{enumerate}

\subsection{Performance on LSMOP Problems and Discussion}
\label{sec:Performance}

\begin{table*}
  \centering
  \caption{Mean Values of SP Metric Obtained by Compared Algorithms Over LSMOP Problems With 3-Objectives}
    \begin{tabular}{c|c|c|c|c|c|c}
    \hline\noalign{\smallskip}
    Problem & Des.  & NSGAIII & RMMEDA & LMEA  & LMOCSO & \textbf{LT-PPM} \\
    \noalign{\smallskip}\hline\noalign{\smallskip}
          & 1000  & 4.37E-01 & 4.36E-01 & 4.72E-01 & 4.75E-01 & \textbf{4.15E-01} \\
    LSMOP1 & 2000  & 4.43E-01 & \textbf{4.28E-01} & 4.33E-01 & 5.16E-01 & 4.93E-01 \\
          & 5000  & 4.35E-01 & 4.32E-01 & 4.19E-01 & 5.33E-01 & \textbf{3.15E-01} \\
    \noalign{\smallskip}\hline\noalign{\smallskip}
          & 1000  & \textbf{1.68E-02} & 4.80E-01 & 5.34E-01 & 6.19E-02 & 3.10E-01 \\
    LSMOP2 & 2000  & \textbf{1.37E-02} & 4.81E-01 & 5.22E-01 & 5.27E-02 & 3.10E-01 \\
          & 5000  & \textbf{1.47E-02} & 5.44E-01 & 5.35E-01 & 5.24E-02 & 3.16E-01 \\
    \noalign{\smallskip}\hline\noalign{\smallskip}
          & 1000  & 8.62E-01 & 9.27E-01 & 1.07E+00 & 6.67E-01 & \textbf{1.76E-01} \\
    LSMOP3 & 2000  & 8.77E-01 & 9.72E-01 & 1.02E+00 & 8.36E-01 & \textbf{2.09E-01} \\
          & 5000  & 9.23E-01 & 9.03E-01 & 1.09E+00 & 9.86E-01 & \textbf{3.66E-01} \\
    \noalign{\smallskip}\hline\noalign{\smallskip}
          & 1000  & \textbf{5.24E-02} & 4.81E-01 & 5.12E-01 & 1.11E-01 & 2.97E-01 \\
    LSMOP4 & 2000  & \textbf{2.63E-02} & 5.13E-01 & 5.22E-01 & 7.39E-02 & 3.10E-01 \\
          & 5000  & \textbf{1.63E-02} & 5.14E-01 & 5.19E-01 & 5.62E-02 & 2.99E-01 \\
    \noalign{\smallskip}\hline\noalign{\smallskip}
          & 1000  & 6.61E-01 & 7.61E-01 & 7.25E-01 & 7.13E-01 & \textbf{4.39E-01} \\
    LSMOP5 & 2000  & 5.75E-01 & 7.04E-01 & 7.74E-01 & 6.58E-01 & \textbf{4.10E-01} \\
          & 5000  & 5.53E-01 & 6.86E-01 & 7.68E-01 & 7.04E-01 & \textbf{3.70E-01} \\
    \noalign{\smallskip}\hline\noalign{\smallskip}
          & 1000  & 9.77E-01 & 1.02E+00 & 1.15E+00 & 1.30E+00 & \textbf{3.08E-01} \\
    LSMOP6 & 2000  & 9.44E-01 & 1.03E+00 & 1.15E+00 & 1.24E+00 & \textbf{2.58E-01} \\
          & 5000  & 9.57E-01 & 1.15E+00 & 1.17E+00 & 9.31E-01 & \textbf{3.07E-01} \\
    \noalign{\smallskip}\hline\noalign{\smallskip}
          & 1000  & 8.59E-01 & 8.73E-01 & 9.46E-01 & 3.57E-01 & \textbf{2.86E-01} \\
    LSMOP7 & 2000  & 5.97E-01 & 9.11E-01 & 9.11E-01 & 1.02E+00 & \textbf{3.13E-01} \\
          & 5000  & 5.21E-01 & 9.37E-01 & 8.99E-01 & 3.91E-01 & \textbf{3.44E-01} \\
    \noalign{\smallskip}\hline\noalign{\smallskip}
          & 1000  & 5.24E-01 & 7.39E-01 & 7.23E-01 & 9.56E-01 & \textbf{3.39E-01} \\
    LSMOP8 & 2000  & 4.54E-01 & 7.53E-01 & 7.63E-01 & 8.58E-01 & \textbf{3.29E-01} \\
          & 5000  & 3.89E-01 & 7.07E-01 & 7.42E-01 & 8.99E-01 & \textbf{3.19E-01} \\
    \noalign{\smallskip}\hline\noalign{\smallskip}
          & 1000  & 8.58E-01 & 7.54E-01 & 9.46E-01 & 8.93E-01 & \textbf{3.66E-01} \\
    LSMOP9 & 2000  & 8.63E-01 & 8.16E-01 & 9.05E-01 & 9.67E-01 & \textbf{3.87E-01} \\
          & 5000  & 8.50E-01 & 8.26E-01 & 9.17E-01 & 9.77E-01 & \textbf{5.32E-01} \\
    \hline\noalign{\smallskip}
    \end{tabular}%
  \label{tab:TableSP}%
\end{table*}%

\begin{table*}
  \centering
  \caption{Mean Values of IGD Metric Obtained by Compared Algorithms Over LSMOP Problems With 3-Objectives}
    \begin{tabular}{c|c|c|c|c|c|c}
    \hline\noalign{\smallskip}
    Problem & Des.  & NSGAIII & RMMEDA & LMEA  & LMOCSO & \textbf{LT-PPM} \\
    \noalign{\smallskip}\hline\noalign{\smallskip}
          & 1000  & 5.4E+00 & 8.8E+00 & 1.1E+01 & 1.3E+00 & \textbf{6.5E-01} \\
    LSMOP1 & 2000  & 7.3E+00 & 7.8E+00 & 1.1E+01 & 1.5E+00 & \textbf{5.9E-01} \\
          & 5000  & 9.8E+00 & 8.6E+00 & 1.1E+01 & 1.4E+00 & \textbf{7.5E-01} \\
    \noalign{\smallskip}\hline\noalign{\smallskip}
          & 1000  & 3.1E-02 & 3.4E-02 & 3.4E-02 & \textbf{2.6E-02} & 3.9E-02 \\
    LSMOP2 & 2000  & 2.0E-02 & 2.5E-02 & 2.4E-02 & \textbf{1.9E-02} & 3.1E-02 \\
          & 5000  & 1.5E-02 & 2.0E-02 & 2.0E-02 & \textbf{1.4E-02} & 2.7E-02 \\
    \noalign{\smallskip}\hline\noalign{\smallskip}
          & 1000  & 1.4E+01 & 1.6E+01 & 5.5E+01 & 9.9E+00 & \textbf{4.9E+00} \\
    LSMOP3 & 2000  & 1.7E+01 & 1.6E+01 & 4.2E+01 & 1.1E+01 & \textbf{4.7E+00} \\
          & 5000  & 1.9E+01 & 1.7E+01 & 2.1E+02 & 1.1E+01 & \textbf{5.1E+00} \\
    \noalign{\smallskip}\hline\noalign{\smallskip}
          & 1000  & 1.1E-01 & 1.1E-01 & 1.1E-01 & \textbf{8.3E-02} & 1.1E-01 \\
    LSMOP4 & 2000  & 5.8E-02 & 6.3E-02 & 6.2E-02 & \textbf{4.8E-02} & 6.4E-02 \\
          & 5000  & 2.8E-02 & 3.2E-02 & 3.2E-02 & \textbf{2.5E-02} & 3.7E-02 \\
    \noalign{\smallskip}\hline\noalign{\smallskip}
          & 1000  & 9.5E+00 & 1.1E+01 & 1.9E+01 & 3.0E+00 & \textbf{6.4E-01} \\
    LSMOP5 & 2000  & 1.2E+01 & 1.2E+01 & 1.9E+01 & 3.4E+00 & \textbf{4.8E-01} \\
          & 5000  & 1.8E+01 & 1.3E+01 & 2.0E+01 & 3.3E+00 & \textbf{7.6E-01} \\
    \noalign{\smallskip}\hline\noalign{\smallskip}
          & 1000  & 3.7E+03 & 7.2E+03 & 3.1E+04 & 3.5E+02 & \textbf{1.9E+00} \\
    LSMOP6 & 2000  & 8.8E+03 & 1.2E+04 & 3.5E+04 & 5.0E+02 & \textbf{1.0E+00} \\
          & 5000  & 1.8E+04 & 5.8E+03 & 3.2E+04 & 5.6E+02 & \textbf{8.7E+00} \\
    \noalign{\smallskip}\hline\noalign{\smallskip}
          & 1000  & 1.1E+00 & 1.1E+00 & 5.4E+02 & \textbf{8.2E-01} & 2.8E+00 \\
    LSMOP7 & 2000  & 1.0E+00 & 1.0E+00 & 1.4E+02 & \textbf{7.2E-01} & 4.0E+00 \\
          & 5000  & 9.7E-01 & 9.7E-01 & 1.2E+02 & \textbf{6.5E-01} & 2.1E+01 \\
    \noalign{\smallskip}\hline\noalign{\smallskip}
          & 1000  & 5.8E-01 & 5.8E-01 & 5.9E-01 & 5.0E-01 & \textbf{2.0E-01} \\
    LSMOP8 & 2000  & 5.6E-01 & 5.8E-01 & 5.6E-01 & 4.9E-01 & \textbf{1.9E-01} \\
          & 5000  & 5.6E-01 & 5.5E-01 & 5.6E-01 & 4.9E-01 & \textbf{1.9E-01} \\
    \noalign{\smallskip}\hline\noalign{\smallskip}
          & 1000  & 2.2E+01 & 4.1E+01 & 1.4E+02 & 2.4E+00 & \textbf{1.4E+00} \\
    LSMOP9 & 2000  & 4.3E+01 & 7.3E+01 & 1.4E+02 & 8.8E+00 & \textbf{1.4E+00} \\
          & 5000  & 6.9E+01 & 8.7E+01 & 1.5E+02 & 4.2E+01 & \textbf{1.4E+00} \\
    \hline\noalign{\smallskip}
    \end{tabular}%
  \label{tab:TableIGD}%
\end{table*}%

Table \ref{tab:TableSP} and Table \ref{tab:TableIGD} show the statistical results of the average SP and IGD values for 10 runs respectively, where the best result on each test instance is shown in bold.

From Table \ref{tab:TableSP} and Table \ref{tab:TableIGD}, it obviously that the performance of the proposed LT-PPM is better than the other five comparative convergence algorithms. Table \ref{tab:TableSP} lists the SP values of the six comparison algorithms. LT-PPM performed best in 20 out of 27 test instances, followed by NSGAIII with 6 best results, and RMMEDA with 1 best result. Specifically, under all decision variable settings, LT-PPM performed better on LSMOP3, LSMOP5, LSMOP6, LSMOP7, LSMOP8 and LSMOP9. In other test instances, LT-PPM is slightly lower than the corresponding best performance algorithm.

In Table \ref{tab:TableIGD}, LT-PPM got 18 of the 27 best results, LMOCSO got 9 best results, and the other algorithms did not get the best results. Specifically, under all decision variable settings, LT-PPM performs better on LSMOP1, LSMOP3, LSMOP5, LSMOP6, LSMOP8 and LSMOP9. LMOCSO achieved better convergence on problems LSMOP2, LSMOP4 and LSMOP7.

The above experimental results show that for most test instances, LT-PPM can obtain a set of solutions with good convergence and diversity. It can be seen that the proposed LT-PPM can obtain competitive performance on the LSMOP test problem, which confirms the effectiveness of LT-PPM in solving the challenging large-scale decision variable problem.

\subsection{Experiment on Time Complexity and Discussion}
\label{sec:TimeComplexity}
\begin{figure}
\centering
\includegraphics[width=0.8 \textwidth]{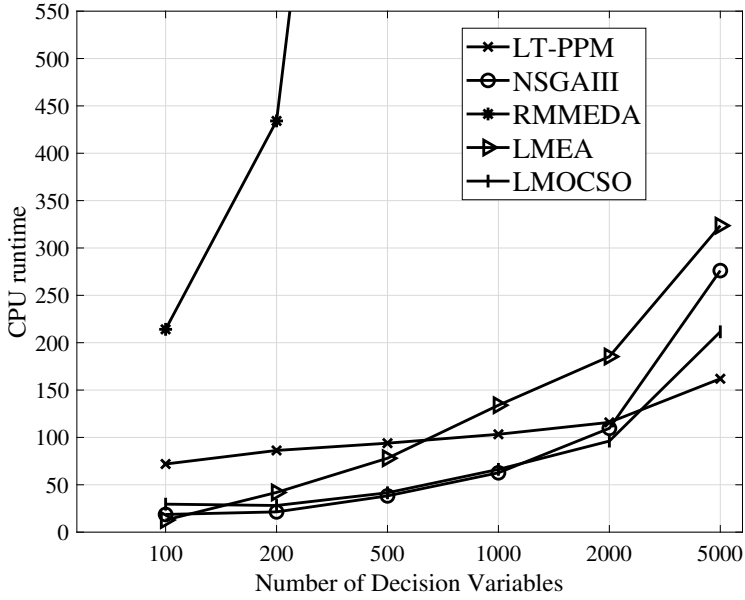}
\caption{Experiment on Time Complexity}
\label{fig:tc}       
\end{figure}
Figure \ref{fig:tc} compares the CPU running time on the LSMOP1 problem of the algorithms in the experiment, where the abscissa is the different decision variables, and the ordinate is the CPU time.  It is worth noting that when the decision variable is greater than 500, the running time of RM-MEDA is greatly increased.

It can be seen from the results that the time consumption of our proposed algorithm is not the lowest when there are fewer decision variables. However, as the decision variables increase, the running time of the proposed algorithm grows slower than other algorithms. When it reaches 5000 dimensions, the time consumption of our algorithm is the lowest.

The results of the experiment prove that our algorithm is not directly related to the decision variables. This allows the algorithm to have better convergence and diversity when solving large-scale multi-objective optimization problems, while also avoiding the impact of large-scale decision variables on computational complexity.

\section{Conclusion and Future Works}
\label{sec:CaFW}
This paper proposes an algorithm, called LT-PPM, for solving LSMOP. The TPM component of the proposed method enables us to effectively use the heuristic information of the individual motion process to predict the POF for next generation. At the same time, the importance sampling technique makes the algorithm independent of the numbers of decision variables, so that the algorithm can explore faster in the large-scale search space. Simultaneously, the importance sampling and the TPM model are used to adjust the sparsity of the population, thereby balancing the development and exploration of the algorithm and escaping from the local optimum.

In LT-PPM algorithm, the general modeling-sampling process is not used. Instead, the Generating-Filtering strategy is adopted, which enables the algorithm to focus on using heuristic information to generate individuals and screen out dominant populations. Our method is compared with several other large-scale multi-objective optimization algorithms on 9 LSMOPs. The experimental results show that LT-PPM has good performance in convergence, distribution and robustness.

This paper presents a preliminary work on the application of trend prediction models to large-scale evolutionary calculations. Future work may take the following possible directions. The method has some details that can be improved, for large-scale problems with complex search spaces, such as constraints and discontinuous Pareto optimal regions, the proposed framework still has a lot of room for improvement. In addition, a wealth of advanced machine learning techniques can stimulate more innovations to solve practical applications with large-scale decision variables.

\begin{acknowledgements}
\textbf{Acknowledgement} This work was supported by the National Natural Science Foundation of China (No.61673328).
\end{acknowledgements}

%
%

\bibliographystyle{unsrt}
\bibliography{LT_PPM}   


\end{document}